\documentclass[runningheads]{llncs}
\usepackage{cite}
\usepackage{url}
\usepackage{siunitx}
\usepackage{subcaption}
\usepackage[T1]{fontenc}
\usepackage{graphicx}
\begin{document}

\title{RF+clust for Leave-One-Problem-Out Performance Prediction}

\author{Ana Nikolikj\inst{1, 2}\orcidID{0000-0002-6983-9627}\and
Carola Doerr\inst{2}\orcidID{0000-0002-4981-3227} \and
Tome Eftimov\inst{1}\orcidID{0000-0001-7330-1902}}

\institute{Computer Systems Department, Jozef Stefan Institute, Ljubljana 1000, Slovenia \and
Jozef Stefan International Postgraduate School, Ljubljana 1000, Slovenia \and Sorbonne Universit\'e, CNRS, LIP6, Paris, France}

\maketitle

\begin{abstract}
Per-instance automated algorithm configuration and selection are gaining significant moments in evolutionary computation in recent years. Two crucial, sometimes implicit, ingredients for these automated machine learning (AutoML) methods are 1) feature-based representations of the problem instances and 2) performance prediction methods that take the features as input to estimate how well a specific algorithm instance will perform on a given problem instance. Non-surprisingly, common machine learning models fail to make predictions for instances whose feature-based representation is underrepresented or not covered in the training data, resulting in poor generalization ability of the models for problems not seen during training.   
In this work, we study leave-one-problem-out (LOPO) performance prediction. We analyze whether standard random forest (RF) model predictions can be improved by calibrating it with a weighted average of performance values obtained by the algorithm on problem instances that are sufficiently close to the problem for which a performance prediction is sought, measured by cosine similarity in feature space. 
While our RF+clust approach obtains more accurate performance prediction for several problems, its predictive power crucially depends on the chosen similarity threshold as well as on the feature portfolio for which the cosine similarity is measured, thereby opening a new angle for feature selection in a zero-shot learning setting, as LOPO is termed in machine learning. 

\keywords{Algorithm Performance Prediction  \and AutoML \and Zero-Shot Learning \and Single-Objective Black-Box Optimization.}
\end{abstract}

\section{Introduction}
Various algorithms for continuous single-objective optimization (SOO) have already been developed and their performance investigated through statistical analyses, in most cases reporting the average performance across a selected set of benchmark problem instances~\cite{stork2020new}. However, the algorithm instance behavior varies substantially depending on the problem instance that is being solved. For this purpose, there is a predictive task known as automated algorithm selection, where the main goal is selecting the best-performing algorithm from a set of algorithm instances for a given problem instance~\cite{malan2014fitness,belkhir2017per,derbel2019new, jankovic2020landscape}. To achieve it, automated algorithm performance prediction is a crucial step that should be done. The automated algorithm performance prediction is tackled as a supervised machine learning problem, such as classification (i.e., predicts whether or not the algorithm solves the instance given some precision) or a regression (i.e., predicts the performance of the algorithm as a real value). To train a supervised machine learning algorithm a set of examples, problem instances described by their landscape features used as input data (i.e., benchmark suite) and algorithm performance achieved on them used as target (i.e., solution precision) are required. Nowadays, if we want to generalize a supervised prediction model to another benchmark suite, whose problems were not involved in the training data, the predictive model performance decreases greatly. This occurs because the training data does not cover some regions of the landscape space involved in the test data, or the model is biased toward some over-represented landscape regions that are present in the training data. This is the reason why most of the studies focus on the Black-box Optimization Benchmarking (BBOB) benchmark suite~\cite{hansen2010real} since it involves several instances from the same problem class involved in the training data which makes it a suitable resource for training an ML-supervised model. However, if we remove the instances from the same problem class from the training data (leave all instances from a single problem out for testing), the performance of the model for automated algorithm performance prediction decreases significantly~\cite{vskvorc2022transfer}. This issue makes all approaches for automated algorithm performance prediction that are developed based on the BBOB benchmark suite difficult to generalize on other benchmark suites such as CEC~\cite{liang2013problem}, Nevergrad~\cite{nevergrad}, etc. since in their definition there is only a single instance per each problem class. 
In practice, an application such as algorithm performance prediction requires making predictions for problem instances whose problem class (i.e., landscape properties) has not been seen previously by the underlying model or we have a leave-one-problem-out (LOPO) learning scenario. 

\textbf{Our contribution.} In this study, we propose a LOPO approach for automated algorithm performance prediction. Our RF+clust approach calibrates a classic random forest (RF) predictive model with a prediction obtained by a similarity relationship method that aggregates the ground algorithm instance performance for the most similar problem instances from the training data. In contrast to a classic KNN approach, we use a similarity threshold to decide which problems are taken into account for the performance prediction. The number of considered `neighbors' can therefore differ between problems. We evaluate our approach on performance data of three differential evolution (DE) variants on the CEC 2014 and the BBOB benchmark suite of the COmparing Continuous Optimizers (COCO) environment ~\cite{hansen2021coco}. We observe better results for RF+clust than for stand-alone RF performance prediction on a number of cases. However, there are also cases when similar landscape representations can lead to different performances of the algorithm, which can affect the prediction of the RF+clust approach. This further points out that in the future we need to focus on finding problem feature representations with sufficient discriminate power that will be also able to capture the performance of the algorithm. 

\textbf{Outline.} The remainder of the paper is organized as follows: Section~\ref{sec:relatedwork} surveys past work on automated algorithm performance prediction. The proposed LOPO regression method is introduced in Section~\ref{sec:zsl}. Section~\ref{sec:experimentaldesign} details the benchmark problem suites and algorithms used for the validation of the proposed approach, the problem landscape features, as well as the machine learning algorithm tuning and evaluation. The results and discussion are provided in Section~\ref{sec:results}. Finally, the conclusions are drawn in Section~\ref{sec:conclusions}.

\section{Related Work}
\label{sec:relatedwork}
Next, we point out some of the works which are addressing the critical issue of generalization over new problem classes.

Bischl et al.~\cite{bischl2012algorithm} consider automated algorithm selection as a cost-sensitive classification task using one-sided Support Vector Machines. Problem instance-specific miss-classification costs are defined, unlike standard classification where all the errors in classification are penalized the same by the algorithm. The predefined miss-classification costs represent external information to aid the learning process. The approach was tested on problem instance feature representations consisting of “cheap” and “expensive” ELA features~\cite{mersmann2011exploratory}, with respect to sample size required for their calculation. It was shown that the model is able to generalize over new instances, however, the prediction error gets worse for new problem classes, when the prediction is based only on the “cheap” ELA features representation. To discover the source of the larger model errors, analysis of the feature space is performed based on the euclidean distances between the problem instances representations. They conclude that the degree of classification performance tends to correlate with the proximity in feature space for the case of using the entire feature set, however, this was not that straightforward for the “cheap” features.

The work~\cite{eftimov2021personalizing} brings to attention the possibility to \emph{personalize} regression models (Decision Tree, Random Forest and Bagging Tree Regression) to specific types of optimization problems. Instead of aiming for a single model that works well across a whole set of possibly diverse problems, the personalized regression approach acknowledges that different models may suite different problem types. 

In~\cite{vskvorc2022transfer} a classification-based algorithm selection approach is evaluated on the COCO benchmark suite~\cite{hansen2010real} and artificially generated problems~\cite{dietrich2022increasing}. The results show that such a model has low generalization power between datasets, and in the leave-one problem-out cross-validation procedure where each problem class was removed once at a time from the same dataset. However, a model trained and tested in a leave-instance-out scenario achieves much higher accuracy. A correlation analysis using the Pearson correlation coefficient was performed for the problem representations based on the “cheap” ELA features, showing that a large number of both, the COCO and the artificial problems are highly correlated within their own set of problems. i.e., the poor generalization is due to the differences between the two data sets in feature space.

The feasibility of a ``per-run'' algorithm selection scheme is investigated in~\cite{kostovska2022per}, based on ELA features that are calculated from the observed trajectory of the algorithm (i.e., the samples the algorithm visits during the optimization procedure). This avoids the usually required additional evaluation of (quasi-)random samples implemented by classic per-instance algorithm selection schemes. Results for the COCO benchmark suite show performance comparable to the per-function virtual best solver. However, these results did not directly generalize to the other benchmark suites used in the experiments, namely the YABBOB suite from the Nevergrad platform~\cite{nevergrad}.

\section{LOPO Algorithm Performance Prediction}
\label{sec:zsl}
 Let us assume a set of benchmark problem instances $P_{t}^i$, $i=1,\dots, n$, which are grouped into training problem classes $P_{t}$, $t=1\dots, m$, and performance data for an algorithm instance $A$ on the selected set of benchmark problem instances. To predict the performance $y_{q}^i$ of the algorithm instance $A$ on a problem instance $P_{q}^i$ from a new problem class $P_{q}$ that is not involved in the training data, we have propose the following LOPO approach:\\ 
1) Represent the selected benchmark problem instances from the $m$ problem classes by calculating the ELA features and linking them to the performance of the algorithm instance after a certain number of function evaluations.\\
2) Train a supervised regression model that uses the ELA features as input data and predicts the algorithm instance performance.\\
3)  For a new problem instance $P_{q}^i$ from a new problem class $P_{q}$ that is not involved in the training data, use its ELA features as input data into the learned model to make the prediction $y_{q}^i$.\\
4) For the new problem instance, select the $k$-nearest problem instances from the training set that are the most similar to the new problem instance based on their landscape features representation. The similarity is measured by a similarity metric, $s$, and the selection is done by defining a prior similarity threshold. We selected cosine similarity as a similarity measure. Finally, all problem instances from the training data that have a similarity greater or equal to the predefined threshold are selected, from which the actual algorithm performance is retrieved $p_{1}, p_{2},\dots, p_{k}$. We need to point out here that the number of the nearest problem instances, $k$, differs for different problem instances, so it is discovered by the selection rule and the predefined threshold.\\
5) The final prediction of the algorithm instance performance on the new problem instance is made by calibrating the actual prediction obtained by the learned model $y_{q}^i$ with the actual algorithm instance performance retrieved for the selected nearest problem instances from the training data. This is performed as an aggregation procedure as follows:
$\widehat{y}_{q}^i$ = $\left(y_{q}^i + F(p_{1}, p_{2},\dots, p_{k})\right)/2$, 
where $F(p_{1}, p_{2},\dots, p_{k})$ is an aggregation function, which can be for example
Weighted mean.
6) If there are no problem instances in the training data to which the new problem instance is similar above the threshold, only the prediction of the model is considered, $y_{q}$.

\section{Experimental Design}
\label{sec:experimentaldesign}
Here, we are going to present all experimental details starting from the data that is involved and the techniques used for the ML learning process.

\textbf{Problem Benchmark Suites.} We evaluate the proposed method by using two of the most currently used benchmark problem suites in the field of numeric single-objective optimization. The first benchmark suite involved in the experiments is the 2014 CEC Special Sessions \& Competitions (CEC 2014) suite. The suite consists of 30 problems where only one instance per problem is available. The problems are provided in dimension 10. The full problem list and descriptions of all the problems are available at \cite{liang2013problem}.
The second problem set is the 24 noiseless single-objective optimization problems from the BBOB collection of the COCO benchmarking platform~\cite{hansen2021coco}. Different problem instances can be derived by transforming the base problem with predefined transformations to both its domain and its objective space, resulting in a set of different instances for each base problem class, that have the same global characteristics. We consider the first five instances of each BBOB problem, resulting in a dataset 120 problem instances. In coherence with the CEC problem suite, the problem dimension $D$ was set to 10.

\begin{figure}[!htb]
\centering
\includegraphics[width=\textwidth]{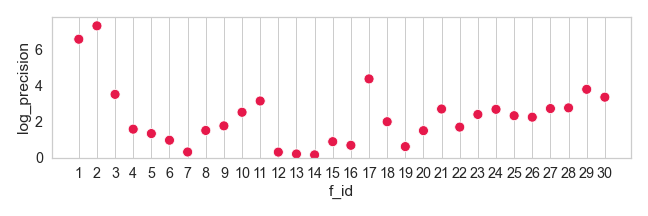}
\caption{DE$1$ solution precision (log-scale) per problem instance on the CEC 2014 suite.}
\label{fig1}
\end{figure}

\begin{figure}[!htb]
\centering
\includegraphics[width=\textwidth]{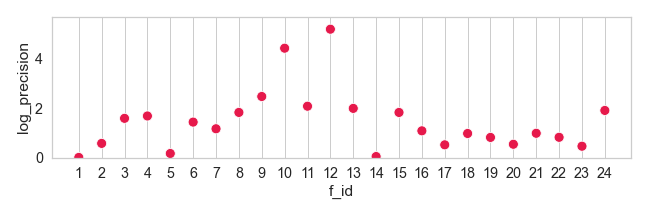}
\caption{DE$1$ solution precision (log-scale) per problem instance on the BBOB suite.}
\label{fig2}
\end{figure}

\textbf{Algorithm Performance Data.} Performance data is collected for three different randomly selected Differential Evolution (DE)~\cite{storn1997differential} configurations, on both the CEC 2014 and BBOB benchmark suites. The DE hyper-parameters are as presented in Table \ref{t:de_configuration}. We indexed the algorithm configurations starting from DE$1$ to DE$3$ for easier notation of the results. DE is an iterative population-based meta-heuristic. The population size of DE is set to equal the problem dimension $D$ ($D=10$ in our study). The three DE configurations were run 30 times on each problem instance, and we extracted the precision after a budget of $500D=5000$ function evaluations. In our study, we consider the median target precision achieved in these 30 runs. 
Following the approach suggested in~\cite{jankovic2020landscape}, we also consider the logarithm (log10) of the median solution precision. This algorithm performance measure estimates the order of magnitude of the distance of the reached solution to the optimum. Figure~\ref{fig1} presents DE$1$ performance (log-scale) obtained per benchmark problem on the CEC 2014, and Figure~\ref{fig2} on the BBOB benchmark suite (aggregated for all problem instances).

\begin{table}[!htb]
\sisetup{round-mode=places}
\caption{Differential Evolution configurations.}
\label{t:de_configuration}
\centering
\begin{tabular}{|l|l| S[round-precision=3]| S[round-precision=3]|}
\hline
$Index$ & $strategy$ & $F$ & $Cr$ \\ \hline
DE$_1$ & Best/3/Bin & 0.533 & 0.809 \\
DE$_2$ & Best/1/Bin & 0.617 & 0.514 \\
DE$_3$ & Rand/Rand/Bin & 0.516 & 0.686 \\
\hline 
\end{tabular}
\end{table}

\textbf{Exploratory Landscape Analysis (ELA).} To create a feature representation that encodes problem properties, the static Exploratory Landscape Analysis (ELA)~\cite{mersmann2011exploratory} features are used. The features are calculated by the evaluation of a sample of candidate solutions generated by systematic sampling of the decision space of the problem. The corresponding fitness values are then fed to different statistical and mathematical methods to calculate the feature values.
The Improved Latin Hypercube Sampling (ILHS)~\cite{xu2017improved} was used as a sampling technique, with a sample size of 800$D$ (8000). In reality, this is a really big sample size, however, we are interested in whether the approach works, so we want to reduce the randomness from the feature computation~\cite{lang2021exploratory}.  For each benchmark problem instance, the calculation of the features was repeated 30 times, as it is a stochastic process and the median value was taken as the final feature value that numerically quantifies some property of the problem. The R package “flacco”~\cite{kerschke2016r} was utilized for their calculation. We selected all the ELA features which are cheap to calculate with regard to sample size, and do not require additional sampling. This way, a total of 64 features were calculated. The selected features are coming from the following groups: classical ELA (y-distribution measures, level-set, meta-model), Dispersion, Information Content, Nearest Better Clustering, and Principal Component Analysis.

\textbf{Regression Models for Algorithm Performance Prediction}.
For the learning process, we considered random forest (RF) regression~\cite{biau2016random}, as it provides promising results for algorithm performance prediction~\cite{jankovic2021impact} and is one of the most commonly used algorithms for algorithm performance prediction studies in evolutionary computation. The RF algorithm was used as implemented by the scikit-learn package~\cite{pedregosa2011scikit} in Python. We have trained single-target regression (STR) models. That is, we have a separate model for predicting the performance of each of the three DE algorithms.

\textbf{ML Model Evaluation.} When splitting the data as described in Section~\ref{sec:zsl} the evaluation of the automated algorithm performance prediction results in leave-one-problem-out fold validation. At each fold, a model was trained using one problem class (including all of its instances) left out for testing, while the remaining are used for training. In order to assess the accuracy of the models, we compute the Mean Absolute Error (MAE). The prediction errors are the absolute distances of the prediction to the true algorithm precision value on the new problem class.

\textbf{Hyper-parameter Tuning for the Regression Models.} The best hyper-parameters are selected for each RF model from the training portion of the fold. The hyper-parameters selected for tuning are $n$ $estimators$ - the number of trees in the random forest;  $max$ $features$ - the number of features used for making the best split; $max$ $depth$ - the maximum depth of the trees, and $min$ $samples$ $split$ - the minimum number of samples required for splitting an internal node in the tree. The ranges of the hyper-parameters have been selected concerning the data set size and the guidelines available in ML to avoid over-fitting. The search spaces of the hyper-parameters for each problem class are presented in our repository [link omitted for review].

\begin{table}[t]
\caption{RF hyper-parameters and the considered search spaces.}
\label{t:rf_hyperparameters_search_scape}
\centering
\begin{tabular}
{l c}
\hline
hyper-parameter & search space\\
\hline
$n$ $estimators$ & \{10, 20, 50, 70\}\\
$max$ $features$ & \{all, sqrt, log2\}\\
$max$ $depth$ &  \{3, 5, 7, 10\} \\
$min$ $samples$ $split$ &  \{2, 5, 7, 10\} \\
\hline 
\end{tabular}
\end{table}

\begin{figure*}[t!]
    \centering
    \begin{subfigure}[t]{0.5\textwidth}
        \centering
        \includegraphics[width=\textwidth]{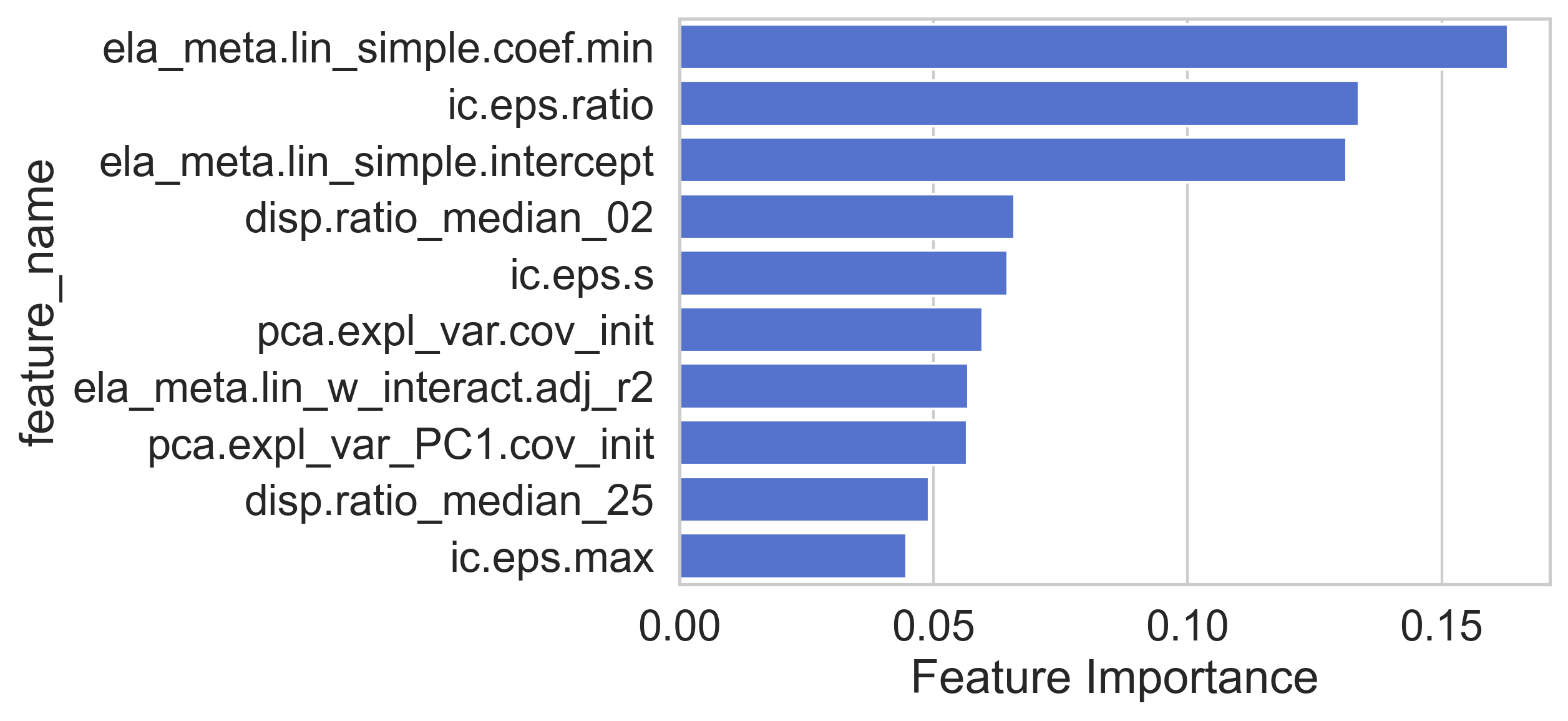}
        \caption{CEC 2014}
    \end{subfigure}%
    ~ 
    \begin{subfigure}[t]{0.5\textwidth}
        \centering
        \includegraphics[width=\textwidth]{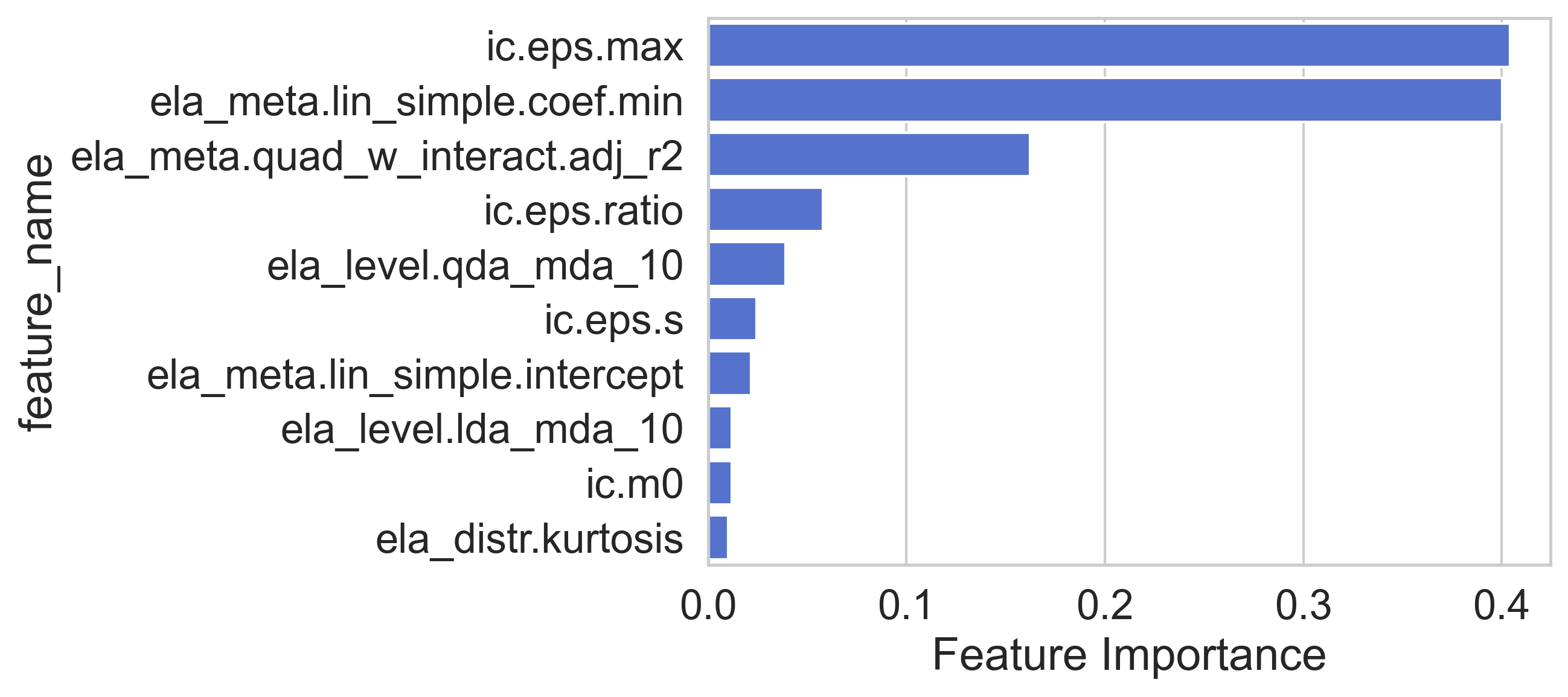}
        \caption{BBOB}
    \end{subfigure}
    \caption{The ten most important ELA features for predicting the performance of DE$1$ on the CEC 2014 and BBOB benchmark suites.}
    \label{fig3}
\end{figure*}

\textbf{Feature Selection.} Taking into consideration the size of the datasets, in a scenario where 30 data instances (CEC 2014 benchmark suite) are available, and 64 features to describe them, we run the risk of overfitting our model.  
Therefore, we have performed feature selection. Since we have a LOPO scenario (i.e., leave all instances for a single problem out), in our case we ended with 30 ML predictive models. To select the top most important features for each model, the SHAP method~\cite{molnar2020interpretable} was utilized. Finally, the importance of the features was summarized across all the models, and the 10 and 30 most important features were used to train the models. These two sizes of feature portfolios were tested in order to compare the decrease in ML model performance (if any) when using different feature portfolios (Table~\ref{t:ml_performance}). Also, the feature portfolio influences the proposed approach as it is based on the pairwise similarity of the features. So different feature portfolios can result in different instances retrieved as similar.
Figure~\ref{fig3} shows the feature portfolio of the 10 most important features for CEC 2014 and BBOB accordingly. 

\begin{table}[tb]
\caption{Mean Absolute Error (mae) obtained by the RF models for predicting the performance of DE$1$ with different feature portfolios on the CEC 2014 benchmark problems}
\label{t:ml_performance}
\centering
\begin{tabular}{ccccc}
\hline
top features &  aggregation &  mae\_train &  mae\_test & \\
\hline
10 &    mean &   0.504464 &  1.279261 &  \\ 
10 &  median &   0.536681 &  0.991274 &  \\ 
30 &    mean &   0.458142 &  1.318326 & \\  
30 &  median &   0.439434 &  0.770265 & \\
64 &    mean &   0.498807 &  1.465239 & \\ 
64 &  median &   0.480457 &  0.931685 & \\
\end{tabular}
\end{table}

\section{Results}
\label{sec:results}
We apply the approach to three random DE configurations and two benchmark suites (CEC 2014 and COCO). Due to space limitations, however, we present here some selected results for algorithm DE$1$ and the CEC 2014 benchmark suite, while other results where similar findings are noticed, are available at [link omitted for review]. 

\begin{figure}[!htb]
\includegraphics[width=\textwidth]{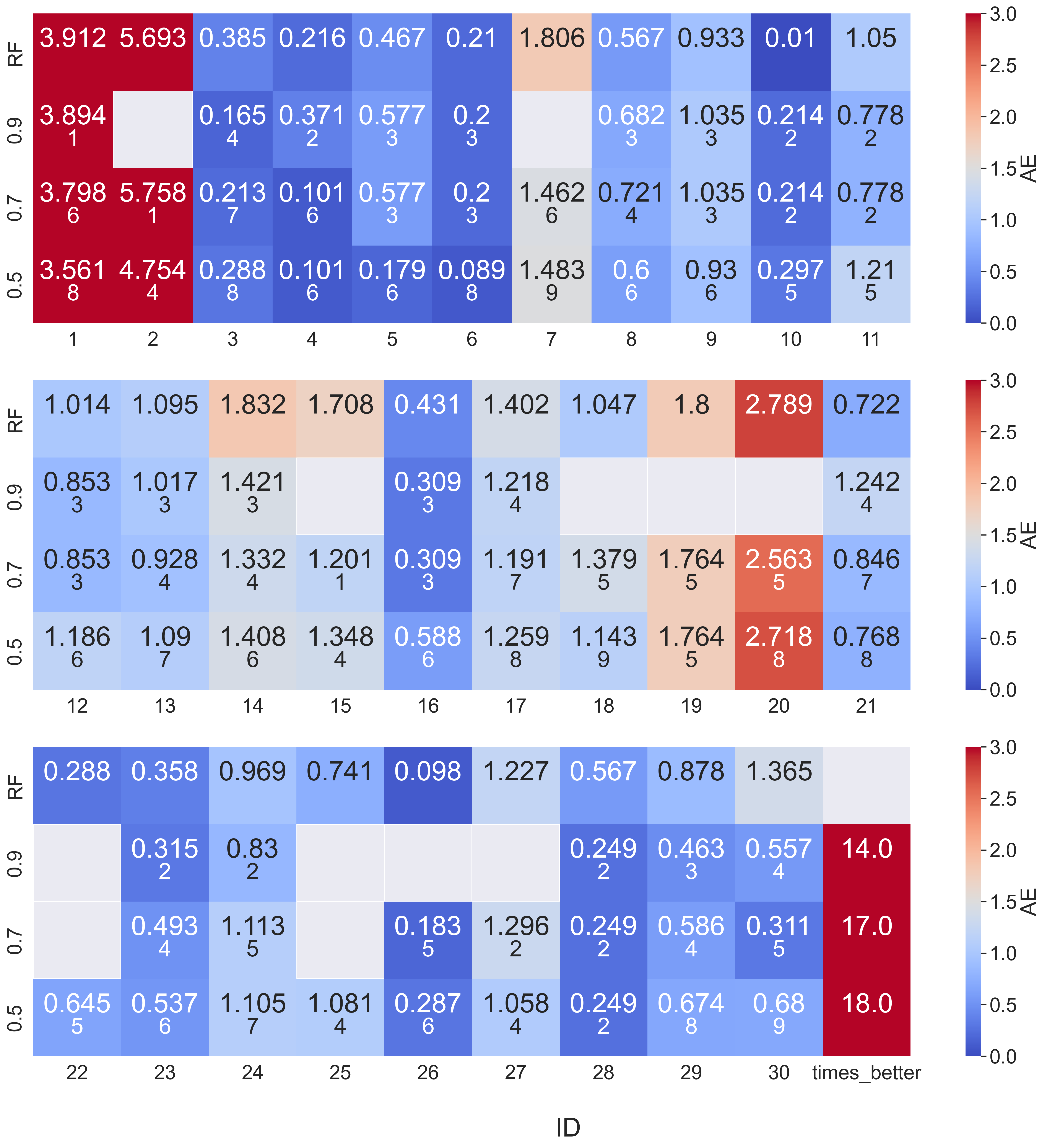}
\caption{The mean absolute error of the RF model and the RF+clust approach using a similarity threshold of .5, .7, and .9, for each problem in the CEC 2014 benchmark suite, for DE1 and the similarity measure based on the 10 most important ELA features.}
\label{fig4}
\end{figure}

In Figure~\ref{fig4} we compare a classical supervised RF model and a RF+clust model in a leave-one-problem-out scenario. Figure~\ref{fig4} shows the prediction errors obtained by a standard RF model trained in the LOPO (corresponding to the "RF" denoted row on the heatmap) and errors of the proposed RF+clust approach (for similarity thresholds of 0.5, 0.7, and 0.9, also with corresponding rows on the heatmap). The predictions were obtained by using a feature portfolio of the ten most important features. Each cell of the heatmap represents the absolute error obtained by the models. The columns represent each problem instance separately, while the last column indicates on how many instances the approach showed lower prediction error).
The numbers under the model error indicate the number of similar instances set above the corresponding threshold that were present in the training of the model. The blank cells in the heatmap are places where the RF+clust approach provides the equal result as the RF model because for those problems we could not find similar problem instances from the training data using the predefined threshold.

\begin{figure}[t]
\centering
\parbox{5cm}{
\includegraphics[width=5cm]{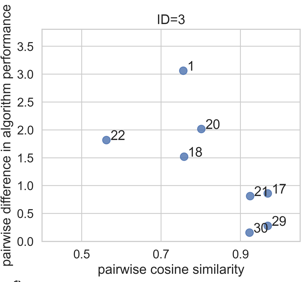}
\caption{Pairwise similarity of the ELA features representation (x-axis) and the pairwise difference in the ground truth performance of DE$1$ (y-axis) for the third problem in CEC 2014.}
\label{fig5}}
\qquad
\begin{minipage}{5cm}
\includegraphics[width=5cm]{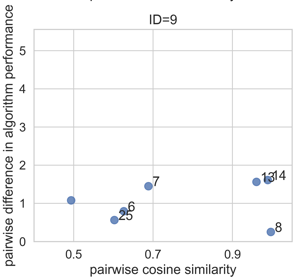}
\caption{Pairwise similarity of the ELA features representation (x-axis) and the pairwise difference in the ground truth performance of DE$1$ (y-axis) for the ninth problem in CEC 2014.}
\label{fig6}
\end{minipage}
\end{figure}

The figure shows that there are problems (1, 3, 6, 11, 12, 13, 14, 16, 17, 23, 24, 28, 29, 30) for which the RF+clust approach shows better predictive results than using a classically trained RF model. We can see that for the high similarity threshold (0.9), calibrating the classical prediction with the ground performance of the optimization algorithm of the retrieved similar problems from the training data decreases the predictive error. To provide an explanation for why this happens, Figure~\ref{fig5} presents the relation between the pairwise similarity of the ELA features representation (x-axis) and the pairwise difference in the ground truth performance of the optimization algorithm (y-axis) for the third problem in the CEC 2014, with the other problems. The heatmap shows that the third problem has four similar problem instances over 0.9 (17, 21, 29, 30), as visible in Figure~\ref{fig5}. In addition, we can see that the difference in ground algorithm performance  of the problem and the similar instances is low, so the algorithm has similar behavior on these problems (see also Figure~\ref{fig1}), and using them for the calibration helps to obtain better predictive errors.

\begin{figure}[t]
\centering
\parbox{5cm}{
\includegraphics[width=5cm]{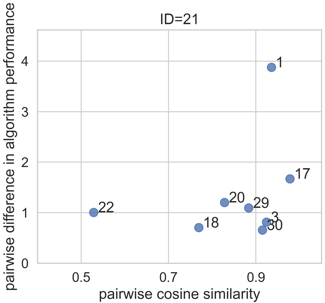}
\caption{Pairwise similarity of the ELA features representation (x-axis) and the pairwise difference in the ground truth in the ground truth performance of DE$1$ (y-axis) for the 21st problem in CEC 2014.}
\label{fig7}}
\qquad
\begin{minipage}{5cm}
\includegraphics[width=5cm]{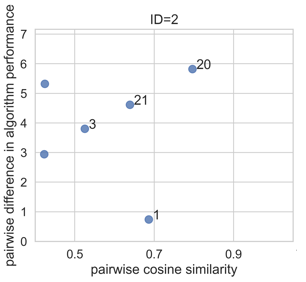}
\caption{Pairwise similarity of the ELA features representation (x-axis) and the pairwise difference in the ground truth in the ground truth performance of DE$1$ (y-axis) for the second problem in CEC 2014.
}
\label{fig8}
\end{minipage}
\end{figure}

There are also problems when the predictions are affected by the RF+clust approach, slightly worse than the prediction obtained from the classical RF model. Figure~\ref{fig6} presents the relation between the pairwise similarity of the ELA features representation (x-axis) and the pairwise difference in the ground truth performance of the optimization algorithm (y-axis) for the ninth problem in the CEC 2014. The heatmap shows that the ninth problem has three similar problem instances (8, 13, 14) according to Figure~\ref{fig6}. Here, we can see that one out of three problems is similarly based on the ELA representation and the algorithm has similar behavior on it. However, on the remaining two problems, we can see that even with high similarity in the landscape space, the difference in algorithm performance is larger in reality (see Figure~\ref{fig1}), so using the performance to calibrate the prediction yields a larger error. A similar scenario happens for the 21st problems, where the similarity is greater or equal to 0.9 but the difference in performance between them is very large (see Figure~\ref{fig7}). This indicates that there are problems for which the ELA features representations are not expressive enough and they could not well describe the problems in such a scenario (i.e., similar ELA landscape representation may not lead to similar algorithm behavior in the performance space). 

There are also problems such as the first and the second that are difficult to be solved by the optimization algorithm (see Figure~\ref{fig1} for ground truth performance). Figure~\ref{fig8} presents the relation between the pairwise similarity of the ELA features representation (x-axis) and the pairwise difference in the ground truth performance of the optimization algorithm (y-axis) for the second problem in the CEC 2014. It is visible that this problem has very few similar instances, does not have similar instances over 0.9 at all, and  also the difference in algorithm performance with similar instances over 0.5 is very large. This is the case where the test problem class is not covered enough by the train, however, even in such scenarios, we can be slightly better in the prediction results.

Looking back at the heatmap (Figure~\ref{fig4}), we can see that when the similarity threshold decreases (i.e., going to 0.5), some of the predictions are slightly worse. This was expected since having a lower threshold in the landscape space does not guarantee capturing similar performance in the performance space (see Figures~\ref{fig6},~\ref{fig7}, and ~\ref{fig8}). The heatmaps presenting the results for the other two DE algorithms lead to similar results and explanations.  

In Table~\ref{t:comparison_cec} we summarize for how many out of the 30 problems the RF+clust approach provides better, worse, or equal predictions (when similar instances are not found, the prediction is not calibrated) than the classical (stand-alone) RF model on the CEC 2014 benchmark suite. 

\begin{table}[tb]
\caption{Number of times the RF+clust is better, equal or worse to stand-alone RF for predicting the performance of DE variants on the CEC 2014 benchmark suite.}
\label{t:comparison_cec}
\centering
\begin{tabular}{lrrrr}
\hline
algorithm name & model& \# better & \# equal &  \# worse\\
\hline
DE1 & 0.9   &         14.0 &        10.0 &         6.0 \\
DE1 & 0.7   &         17.0 &         2.0 &        11.0 \\
DE1 & 0.5   &         18.0 &         0.0 &        12.0 \\
\hline

DE2 & 0.9   &         13.0 &         5.0 &        12.0 \\
DE2 & 0.7   &         18.0 &         0.0 &        12.0 \\
DE2 & 0.5   &         21.0 &         0.0 &         9.0 \\
\hline

DE3 & 0.9   &          8.0 &        15.0 &         7.0 \\
DE3 & 0.7   &         14.0 &         2.0 &        14.0 \\
DE3 & 0.5   &         16.0 &         0.0 &        14.0 \\
\hline
\end{tabular}
\end{table}

\begin{figure}[!htb]
\includegraphics[width=0.95\textwidth]{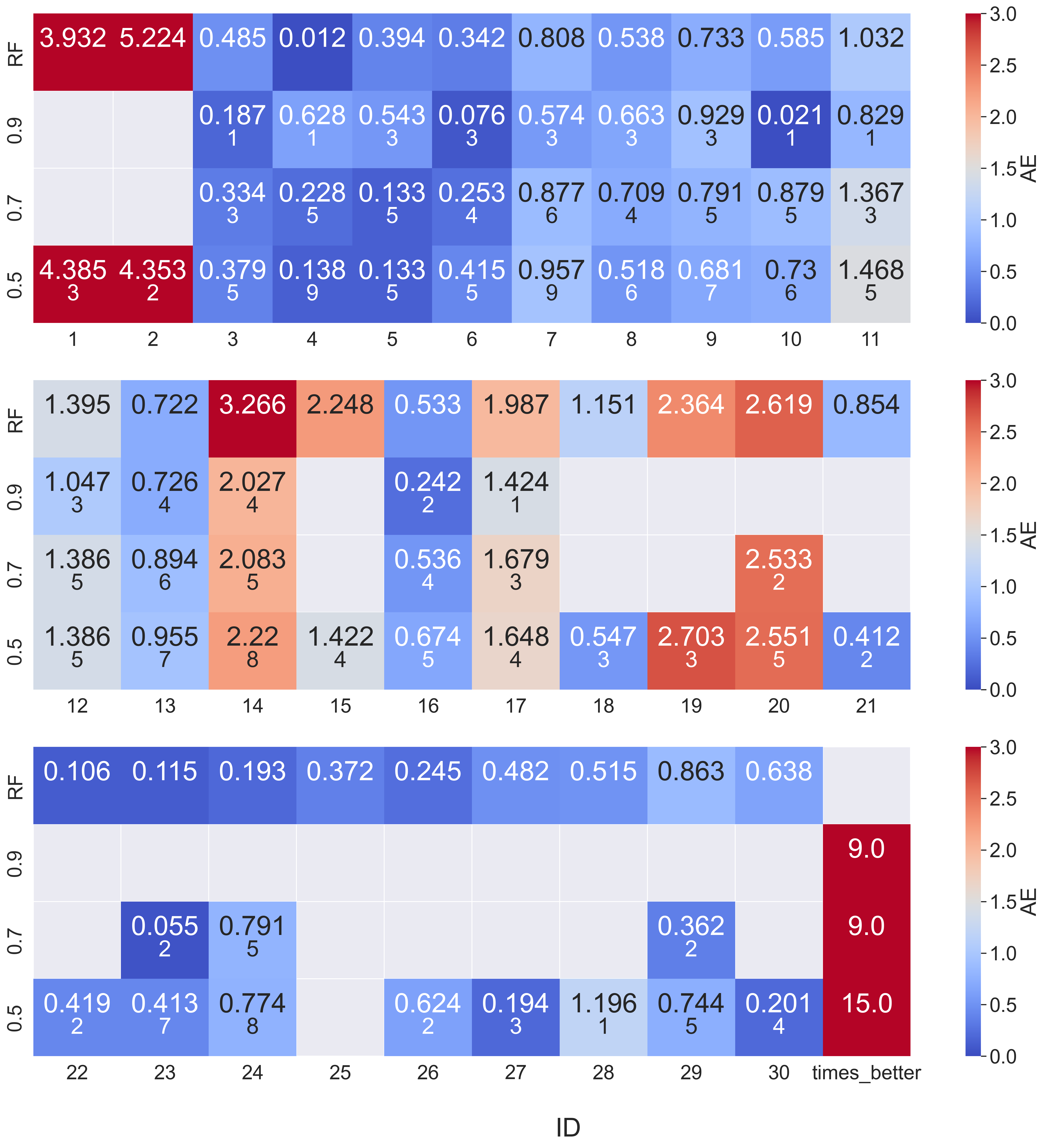}
\caption{The mean absolute error of the RF model and the RF+clust approach using a similarity threshold of .5, .7, and .9, for each problem in the CEC 2014 benchmark suite, for DE1 and the similarity measure based on the 30 most important ELA features.}
\label{fig:30_features}
\end{figure}

To investigate the sensitivity with a different feature portfolio, we repeated the experiments by selecting the top 30 most important features for prediction the performance of DE1 on the CEC 2014 benchmark suite. Figure~\ref{fig:30_features} presents the absolute errors of the RF and RF+clust approaches obtained for each problem of the CEC 2014. Focusing on similarity threshold of 0.9, the results show that the RF+clust approach provides better predictions (i.e., improvements) than the classical RF model for nine problems, worse predictions for five problems, and equal for 16 problems. We need to point out here that increasing the number of the top most important features from 10 to 30, it also affect the similarity of the problem instances. From the heatmap is visible that now we are not able to detect similar problems for some of the problems (e.g., 1, 21, 23, 24, 28, 29, 30) for which we were able to detect similar problems above 0.9 when the feature portfolio of the 10 most important features is used. This further opens new angle for feature selection that will have discriminate power to capture also differences that happen in the performance space. 

\begin{figure}[!htb]
\includegraphics[width=\textwidth]{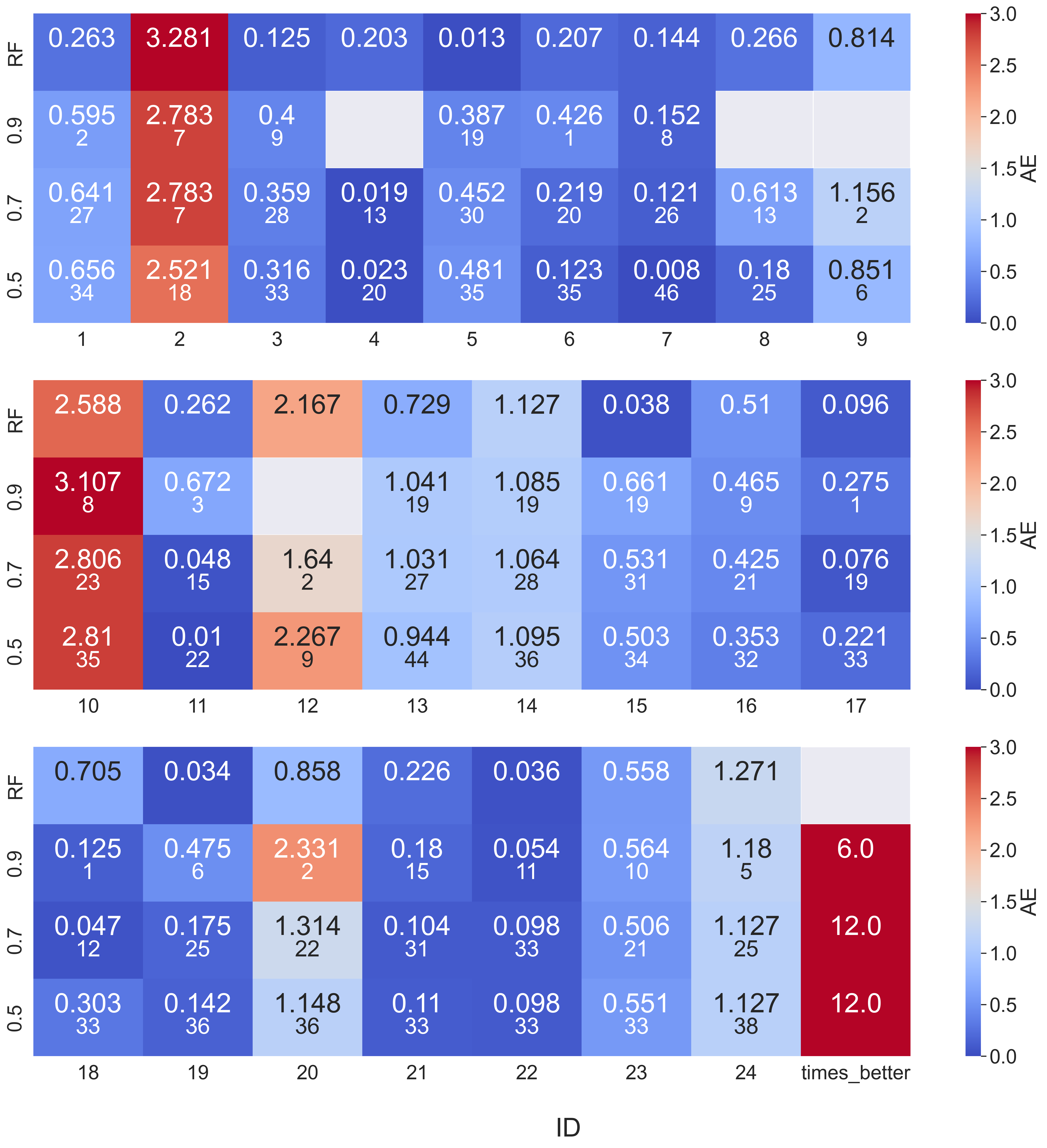}
\caption{The mean absolute error of the RF model and the RF+clust approach using a similarity threshold of .5, .7 and .9, for the first instance of each problem in the BBOB benchmark suite, for DE1 and the ten most important features.}
\label{fig9}
\end{figure}

In addition to the results obtained on the CEC 2014 benchmark suite, Figure~\ref{fig9} presents the prediction results obtained for the DE1 algorithm for the first instance of each COCO problem. We selected only one instance here, for visualization purposes (our overall setting remains LOPO, i.e., we omit \emph{all} instances of the left-out problem, and we use data from the five instances of the other problems for training. This also explains why the number of similar problem instances are larger in Figure~\ref{fig9} compared to those for the CEC benchmark presented in Figure~\ref{fig4}). The results are in line with those obtained for the CEC 2014 benchmark. We point out that the top 10 most important features to train the prediction model differ from those selected on the CEC 2014 benchmark; see Figure~\ref{fig3} for details. An indirect outcome of this study is that these two benchmark suites are different, which also supports previously published findings~\cite{vskvorc2020understanding}. 

\section{Conclusions}
\label{sec:conclusions}
In this study, we have proposed leave-one-problem-out (LOPO) for algorithm performance prediction. The idea behind the approach is to predict the performance of an optimization algorithm by using a supervised ML model on a problem that is not presented in the training data. First, a model is learned from a feature landscape representation of the problem instances from the train problem classes, and a prediction is made for an instance from a new problem class in a supervised manner. Second, based on the similarity relationship between the problem classes based on their feature landscape representation, the prediction for the new problem instance is calibrated by applying an aggregation procedure over the algorithm performance of the $k$-nearest problem instances from the training data.

The results performed on the CEC 2014 benchmark suite showed promising results and explanations about the strengths and weaknesses of the proposed approach. Better results are achieved for problems for which their landscape feature representation is similar to other problems from the training data and the algorithm behaves similarly on those problems. However, there are also problems for which the proposed approach can lead to slightly worse prediction results. This happens for problems for which the landscape feature representation leads to finding similar problems from the training data, however, the performance of the algorithm significantly differs. Such a result indicates that we need to find an expressive enough landscape feature representation that correlates with the algorithm's performance. Also, there are problems for which there are no similar problems in the training data, which further indicates that we need to enrich the data that is used in ML setup with new problems (e.g., merging different benchmark suites or using artificially problem generators~\cite{dietrich2022increasing}) by taking care that all landscape spaces approximate a uniform distribution in the problem space.

In the future, we are planning to test the approach on a more comprehensive algorithm portfolio. Next, instead of exploratory landscape features calculated by a global sampling, we are planning to calculate them using the trajectory data that was observed by the algorithm during the run (i.e., to capture also information about the algorithm behavior)~\cite{kostovska2022per}. We are also going to try different problem feature representations such as topological data analysis~\cite{petelin2022tla}. Last but not least, we are planning to merge different benchmark suites to select representative problem instances~\cite{cenikj2022selector, eftimov2022less} that will allow us to represent all possible landscape spaces from the problem space with the same number of problems that will further help the LOPO approach to have better prediction results.

\small{
 \textbf{Acknowledgments.} 
The authors acknowledge the support of the Slovenian Research Agency through program grant P2-0098, project grants N2-0239 and J2-4460, and a bilateral project between Slovenia and France grant No. BI-FR/23-24-PROTEUS-001 (PR-12040). Our work is also supported by ANR-22-ERCS-0003-01 project VARIATION.}
\bibliographystyle{splncs04}
\bibliography{References}

\end{document}